\documentclass[a4paper]{article}
\usepackage{hyperref}
\usepackage{url}
\usepackage{mystylefile}
\title{Direct Density-Derivative Estimation
and Its Application in KL-Divergence Approximation}
\author{Hiroaki Sasaki \\ sasaki@sg.cs.titech.ac.jp \\ Graduate School
of Information Science and Engineering, \\ Tokyo Institute of
Technology, Tokyo, Japan \\\vspace{-1mm} \\ 
Yung-Kyun Noh \\
nohyung@kaist.ac.kr \\
Department of Computer Science, \\
KAIST, Daejeon, Rep. of Korea
\\ \vspace{-1mm} \\ 
Masashi Sugiyama \\ 
sugi@cs.titech.ac.jp \\
Graduate School of Information Science and
Engineering, \\ Tokyo Institute of Technology, Tokyo, Japan}
\date{}

\begin{document}
\maketitle
\begin{abstract}
 Estimation of density derivatives is a versatile tool in statistical
 data analysis. A naive approach is to first estimate the density and
 then compute its derivative.  However, such a two-step approach does
 not work well because a good density estimator does not necessarily
 mean a good density-derivative estimator.  In this paper, we give a
 direct method to approximate the density derivative without estimating
 the density itself.  Our proposed estimator allows analytic and
 computationally efficient approximation of multi-dimensional high-order
 density derivatives, with the ability that all hyper-parameters can be
 chosen objectively by cross-validation.  We further show that the
 proposed density-derivative estimator is useful in improving the
 accuracy of non-parametric KL-divergence estimation via metric
 learning. The practical superiority of the proposed method is
 experimentally demonstrated in change detection and feature selection.
\end{abstract}
 \section{Introduction}
 Derivatives of probability density functions play key roles in various
 statistical data analysis. For example:
 \begin{itemize}
 \item \emph{Mean shift} clustering seeks modes of the data
       densities~\cite{fukunaga1975estimation,cheng1995mean,comaniciu2002mean,sasaki2014clustering},
       where the first-order density derivative is the key ingredient.
\item The optimal bandwidth of kernel density estimation
      depends on the second-order density derivative~\cite{silverman1986density}.
\item The bias of nearest-neighbor Kullback-Leibler divergence
      estimation is governed by the second-order density derivative~\cite{noh2014bias}.
\item More applications in statistical problems such as
  regression, Fisher information estimation, parameter estimation,
  and hypothesis testing are discussed in~\cite{singh1977applications}.
\end{itemize}
Given such a wide range of applications, accurately estimating the
density derivatives from data has been a challenging research topic in
statistics and machine learning.
 
A naive approach to density-derivative estimation from samples
$\{x_i\}_{i=1}^n$ following probability density $p(x)$ on $\R{}$ is to perform density
estimation and then compute its derivatives.  For example, suppose that
kernel density estimation (KDE) is used for density estimation:
\begin{align*}
  \widehat{p}(x)\propto
\sum_{i=1}^n K\left(\frac{x-x_i}{h}\right),
\end{align*}
where $K$ is a kernel function (such as the Gaussian kernel) and $h>0$
is the bandwidth.  Then the first-order density derivative is estimated
as follows \cite{bhattacharya1967estimation, schuster1969estimation}:
\begin{align*}
 \widehat{p}'(x)\propto
 \sum_{i=1}^n K'\left(\frac{x-x_i}{h}\right).
\end{align*}
A cross-validation method for selecting the bandwidth $h$ was proposed
in \cite{hardle1990bandwidth}.  However, since a good density estimator
is not always a good density-derivative estimator, this approach is not
necessarily reliable; this problem becomes more critical if higher-order
density derivatives are estimated:
\begin{align*}
 \widehat{p}^{(j)}(x)\propto
 \sum_{i=1}^n K^{(j)}\left(\frac{x-x_i}{h}\right).
\end{align*}

A more direct approach of performing kernel density estimation for
density derivatives was proposed \cite{singh1977improvement}:
\begin{align*}
 \widehat{p}^{(j)}(x)\propto
 \sum_{i=1}^n K\left(\frac{x-x_i}{h}\right).
\end{align*}
However, this method suffers the bandwidth selection problem
because the optimal bandwidth depends on higher-order derivatives than
the estimated one \cite{singh1981exact}.

In this paper, we propose a novel density-derivative estimator
which finds the minimizer of the mean integrated square error (MISE)
to the true density-derivative. 
The proposed method, which we call \emph{MISE for derivatives} (MISED),
possesses various useful properties:
\begin{itemize}
 \item Density derivatives are directly estimated without going through
       density estimation.
 \item The solution can be computed analytically and efficiently.
 \item All tuning parameters can be objectively optimized by cross-validation.
 \item Multi-dimensional density derivatives can be directly estimated.
 \item Higher-order density derivatives can be directly estimated.
\end{itemize}
Through experiments on change detection and feature selection,
we demonstrate the usefulness of the proposed MISED method.
\section{Direct Density-Derivative Estimation}
In this section, we describe our proposed MISED method.
\subsection{Problem Formulation}
Suppose that independent and identically
 distributed samples $\mathcal{X}=\{\vector{x}_i\}_{i=1}^n$ from unknown
 density $p(\vector{x})$ on $\R{d}$ are available. Our goal is to
 estimate the $k$-th order (partial) derivative of $p(\vector{x})$,
 \begin{align}
  p_{k,\vector{j}}(\vector{x})=\frac{\partial^k}{\partial
  x_1^{j_1}\partial x_2^{j_2}\dots\partial x_d^{j_d}} p(\vector{x}),
  \label{defi:der}
 \end{align}
 where $j_1+j_2+\dots+j_d=k$ for $j_i\in\{0,1,\dots,k\}$ and
 $\vector{j}=(j_1,j_2,\dots,j_d)$. When $k=1$ (or $k=2$),
 $p_{k,\vector{j}}(\vector{x})$ corresponds to a single element in the
 gradient vector (or the Hessian matrix) of $p(\vector{x})$.

  \subsection{MISE for Density Derivatives}
  Let $g_{k,\vector{j}}(\vector{x})$ be a model of
  $p_{k,\vector{j}}(\vector{x})$ (its specific form will be introduced
  later).  We learn $g_{k,\vector{j}}(\vector{x})$ to minimize the MISE
  to $p_{k,\vector{j}}(\vector{x})$:
  \begin{align}
   J_{\vector{j}}(g_{k,\vector{j}})&=\int
   \left\{g_{k,\vector{j}}(\vector{x})-p_{k,\vector{j}}(\vector{x})\right\}^2
   \mathrm{d}\vector{x} \nonumber \\ &=\int
   \left\{g_{k,\vector{j}}(\vector{x})\right\}^2\mathrm{d}\vector{x} -2\int
   g_{k,\vector{j}}(\vector{x})p_{k,\vector{j}}(\vector{x})\mathrm{d}\vector{x}
   +\underbrace{\int
   \left\{p_{k,\vector{j}}(\vector{x})\right\}^2\mathrm{d}\vector{x}}_{C},
   \label{obj}
  \end{align}
  where $C$ is constant irrelevant to $g_{k,\vector{j}}(\vector{x})$ and
  thus can be safely ignored.

  The first term in (\ref{obj}) is accessible since
  $g_{k,\vector{j}}(\vector{x})$ is a model specified by the user.  The
  second term in (\ref{obj}) seems intractable at a glance, but
  \emph{integration by parts} allows us to transform it as
  \begin{align*}
   \int
   g_{k,\vector{j}}(\vector{x})p_{k,\vector{j}}(\vector{x})\mathrm{d}\vector{x}
   &= \int g_{k,\vector{j}}(\vector{x}) \frac{\partial^k}{\partial
   x_1^{j_1}\partial x_2^{j_2}\dots\partial x_d^{j_d}} p(\vector{x})
   \mathrm{d}\vector{x}, \\ &= \int \left[g_{k,\vector{j}}(\vector{x})
   \frac{\partial^{k-1}}{\partial x_1^{j_1-1}\partial
   x_2^{j_2}\dots\partial x_d^{j_d}}
   p(\vector{x})\right]_{x_1=-\infty}^{x_1=\infty}\mathrm{d}\vector{x}_{\setminus
   x_1}\\ &\phantom{=}-\int \frac{\partial}{\partial
   x_1}g_{k,\vector{j}}(\vector{x}) \frac{\partial^{k-1}}{\partial
   x_1^{j_1-1}\partial x_2^{j_2}\dots\partial x_d^{j_d}}
   p(\vector{x})\mathrm{d}\vector{x},
  \end{align*}
  where $\mathrm{d}\vector{x}_{\setminus x_1}$ denotes the integration
  except for $x_1$. The first term in the last equation vanishes under a
  mild assumption on the tails of $g_{k,\vector{j}}(\vector{x})$ and
  $\frac{\partial^{k-1}}{\partial x_1^{j_1-1}\partial
  x_2^{j_2}\dots\partial x_d^{j_d}} p(\vector{x})$.  By repeatedly
  applying integration by parts $k$ times, we arrive at
  \begin{align*}
   J_{\vector{j}}(g_{k,\vector{j}})&=\int
   \left\{g_{k,\vector{j}}(\vector{x})\right\}^2\mathrm{d}\vector{x} -2(-1)^k\int
   \left\{\frac{\partial^k}{\partial x_1^{j_1}\partial
   x_2^{j_2}\dots\partial x_d^{j_d}}
   g_{k,\vector{j}}(\vector{x})\right\} p(\vector{x})\mathrm{d}\vector{x}+C.
  \end{align*}
  Ignoring the constant $C$ and approximating the expectation by the
  sample average gives
 \begin{align}
  \int \left\{g_{k,\vector{j}}(\vector{x})\right\}^2\mathrm{d}\vector{x}
  -\frac{2(-1)^k}{n} \sum_{i=1}^n \frac{\partial^k}{\partial
  x_1^{j_1}\partial x_2^{j_2}\dots\partial x_d^{j_d}}
  g_{k,\vector{j}}(\vector{x}_i).  \label{MISE-accessible}
 \end{align}
  \subsection{Analytic Solution for Gaussian Kernels}
  As a density-derivative model $g_{k,\vector{j}}$, we use the Gaussian
  kernel model\footnote{ If $n$ is too large, we may only use a subset
  of data samples as kernel centers.  }:
  \begin{align*}
   g_{k,\vector{j}}(\vector{x})&= \sum_{i=1}^n\theta_{\vector{j},i}
   \underbrace{\exp\left(-\frac{\|\vector{x}-\vector{x}_i\|^2}{2\sigma^2}\right)}_{\psi_i(x)}
   =\vector{\theta}_{\vector{j}}^{\top}\vector{\psi}(x),
 \end{align*}
 for which the $k$-th derivative is given by
  \begin{align*}
   \frac{\partial^k}{\partial x_1^{j_1}\partial x_2^{j_2}\dots\partial
   x_d^{j_d}} g_{k,\vector{j}}(\vector{x})
   &=\sum_{i=1}^n\theta_{\vector{j},i}
   \underbrace{\frac{\partial^k}{\partial x_1^{j_1}\partial
   x_2^{j_2}\dots\partial x_d^{j_d}}
   \exp\left(-\frac{\|\vector{x}-\vector{x}_i\|^2}{2\sigma^2}\right)}_{\varphi_{\vector{j},i}(\vector{x})}
   =\vector{\theta}_{\vector{j}}^{\top}\vector{\varphi}_{\vector{j}}(x).
  \end{align*}
  Substituting these formulas into the objective function
  \eqref{MISE-accessible} and adding the $\ell_2$-regularizer, we 
  obtain a practical objective function:
  \begin{align}
   \widetilde{J}_{\vector{j}} (\vector{\theta_j})=
   \vector{\theta}_{\vector{j}}^{\top}\mathbf{G}\vector{\theta_j}
   -2(-1)^{k}\vector{\theta_j}^{\top}\vector{h_j}
   +\lambda\vector{\theta_j}^{\top}\vector{\theta_j}, \label{sample}
  \end{align}
  where 
  \begin{align*}
   [\mathbf{G}]_{ij}&=\int\psi_i(\vector{x})\psi_j(\vector{x})\mathrm{d}\vector{x}
   =(\pi\sigma^2)^{d/2}
   \exp\left(-\frac{\|\vector{x}_i-\vector{x}_j\|^2}{4\sigma^2}\right)
   \quad\mbox{and}\quad \vector{h}_{\vector{j}}
   =\frac{1}{n}\sum_{i=1}^n\vector{\varphi}_{\vector{j}}(\vector{x}_i).
  \end{align*}
  The minimizer of (\ref{sample}) is given analytically as
  \begin{align}
   \widehat{\vector{\theta}}_{\vector{j}}=
   \arg\min_{\vector{\theta_j}}\widetilde{J}_{\vector{j}}(\vector{\theta_j})
   =(-1)^{k}\left(\mathbf{G}^{-1}+\lambda\I\right)\vector{h}_{\vector{j}},
  \end{align}
 where $\I$ denotes the identity matrix.  Finally, a density-derivative
 estimator is obtained as
 \begin{align*}
  \widehat{g}_{k,\vector{j}}(\vector{x})&=
  \widehat{\vector{\theta}}_{\vector{j}}^{\top}\vector{\psi}(x).
 \end{align*}

 We call this method the \emph{mean integrated square error for
 derivatives} (MISED) estimator, which can be regarded as an extension
 of \emph{score matching} for density estimation
 \cite{hyvarinen2005estimation}, \emph{least-squares density-difference}
 for density-difference estimation
 \cite{IEEE-PAMI:Kim+Scott:2010,NC:Sugiyama+etal:2013}, and
 \emph{least-squares log-density gradients} for log-density gradient
 estimation \cite{sasaki2014clustering} to higher-order derivatives.
  \subsection{Model Selection by Cross-Validation}
  \label{ssec:CV}
  The performance of the MISED method depends on the choice of
  model parameters (the Gaussian width
  $\sigma$ and the regularization $\lambda$ in the current setup).
  Below, we describe a method
  to optimize the model by cross-validation,
  which essentially follows the same line as \cite{hardle1990bandwidth}
  for kernel density estimation:
  \begin{enumerate}
   \item Divide the sample $\mathcal{X}=\{\vector{x}_i\}_{i=1}^{n}$ into
	 $T$ disjoint subsets $\{\mathcal{X}_t\}_{t=1}^{T}$. 
	 
   \item The estimator $\widehat{g}^{(t)}_{k,\vector{j}}(\vector{x})$ is
	 obtained using $\mathcal{X}\setminus\mathcal{X}_t$,
         and then the hold-out MISE to
	 $\mathcal{X}_t$ is computed as
	 \begin{align}
\text{CV}(t)&=
	  \int \widehat{g}^{(t)}_{k,\vector{j}}(\vector{x})\mathrm{d}\vector{x}
	  -\frac{2(-1)^{k}}{|\mathcal{X}_t|}
	  \sum_{\vector{x}\in\mathcal{X}_t} \frac{\partial^k}{\partial
	  x_1^{j_1}\partial x_2^{j_2}\cdots\partial x_d^{j_d}}
	  \widehat{g}^{(t)}_{k,\vector{j}}(\vector{x}), \label{CV}
	 \end{align} 
	 where $|\mathcal{X}_t|$ denotes the number of elements in
	 $\mathcal{X}_t$.
	 
   \item The model that minimizes
	 $\text{CV}=\frac{1}{T}\sum_{t=1}^T\text{CV}(t)$ is chosen.
  \end{enumerate}

\subsection{Numerical Examples}
\begin{figure}[!t]
 \centering \includegraphics[width=\textwidth]{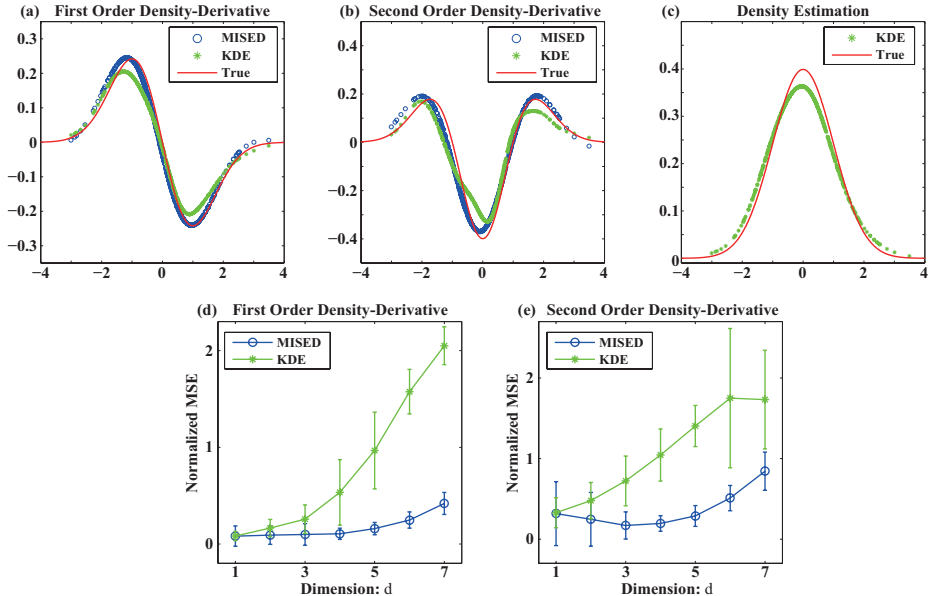}
 \caption{\label{fig:DerEsti} Estimation of the derivatives of the
 standard normal distribution.  (a) and (b): First-order and
 second-order derivative estimation.  (c): Density estimation (only by
 KDE). (d) and (e): Normalized mean squared error (MSE) for first-order
 and second-order derivative estimation as functions of the
 dimensionality of data.}
\end{figure}
Let us illustrate the behavior of MISED using $n=500$ samples drawn from
the standard normal distribution.  The Gaussian bandwidth $\sigma$ and
the regularization parameter $\lambda$ included in MISED are chosen by
5-fold cross-validation from the nine candidates,
$\sigma\in\left\{10^{-0.3}, 10^{-0.1375},\dots,10^{1}\right\}$ and
$\lambda\in\{10^{-1},10^{-0.75},\ldots,10^{1}\}$. For comparison, we
also test the Gaussian KDE where the Gaussian bandwidth $h$ is also
chosen from the same candidate values by 5-fold cross-validation to
minimize the hold-out MISE (\ref{CV}).

Figures~\ref{fig:DerEsti} (a) and (b) depict the estimation results of
the first-order and second-order density derivatives.  The derivatives
estimated by KDE are less accurate than MISED in particular for the
second-order derivative, although the density itself is reasonably
approximated as shown in Figure~\ref{fig:DerEsti} (c).  This result
substantiates that a good density estimator is not necessary a good
density-derivative estimator.

Next, we evaluate how the performance is affected when the
dimensionality of the standard normal distribution is increased.  We use
the common $\sigma$ (and $\lambda$ for MISED) for all $\vector{j}$ and
the summation of the hold-out MISE for all $\vector{j}$ is used as the
cross-validation score.  Figures~\ref{fig:DerEsti} (d) and (e) show that
the increase of the normalized mean squared errors (MSE)\footnote{The
normalized MSE in this paper is defined by
$\frac{\frac{1}{n}\sum_{i=1}^n\sum_{\vector{j}}
(\widehat{g}_{k,\vector{j}}(\vector{x_i})-p_{k,\vector{j}}(\vector{x}_i))^2}{\sqrt{\frac{1}{n}\sum_{i=1}^n\sum_{\vector{j}}
\widehat{g}_{k,\vector{j}}(\vector{x_i})^2\frac{1}{n}\sum_{i=1}^n\sum_{\vector{j}}
p_{k,\vector{j}}(\vector{x_i})^2}}$.} for the MISED method is much
milder than that for KDE, illustrating the high reliability of MISED in
high-dimensional problems.

\section{Application to Kullback-Leibler (KL) Divergence Approximation}
\label{ssec:KL} 
In this section, we apply density-derivative estimation to KL-divergence
approximation.
\subsection{Nearest-Neighbor KL-Divergence Approximation}
The KL-divergence from one density $p_1(\vector{x})$ to another density
$p_2(\vector{x})$, defined as
\begin{align*}
\mathrm{KL}(p_1\|p_2)=\int p_1(\vector{x})
\log\frac{p_1(\vector{x})}{p_2(\vector{x})}\mathrm{d}\vector{x},
\end{align*}
is useful for various purposes such as two-sample homogeneity testing
\cite{IEEE-IT:Kanamori+etal:2012}, feature selection \cite{Gavin09ANew},
and change detection \cite{NN:Liu+etal:2013}.  Here, we consider the
KL-divergence approximator based on \emph{nearest-neighbor density
estimation} (NNDE) \cite{Wang06anearest-neighbor} from two sets of
samples $\mathcal{X}_1=\{\vector{x}_i\}_{i=1}^{n_1}$ and
$\mathcal{X}_2=\{\vector{x}_i\}_{i=n_1+1}^{n_1+n_2}$ following
$p_1(\vector{x})$ and $p_2(\vector{x})$ on $\R{d}$:
\begin{align*}
\widehat{\mathrm{KL}}(p_1\|p_2)=\frac{1}{n_1}\sum_{i=1}^{n_1}
\log\frac{(n_1-1)\mathrm{dist}_1(\vector{x}_i)^d}{n_2\mathrm{dist}_2(\vector{x}_i)^d},
\end{align*}
where $\mathrm{dist}_1(\vector{x})$ and $\mathrm{dist}_2(\vector{x})$
denote the distance from $\vector{x}$ to the nearest samples in
$\mathcal{X}_1$ and $\mathcal{X}_2$, respectively.

\subsection{Metric Learning for NNDE-Based KL-Divergence Approximation}
Although the KL-divergence itself is metric-invariant, the NNDE-based
KL-divergence approximator is metric-dependent.  Indeed, it was shown in
\cite{noh2014bias} that the bias of the NNDE-based KL-divergence
approximator at $\vector{x}$ is approximately proportional to
\begin{align*}
\frac{\tr(\nabla\nabla p_1)}{((n_1-1)p_1)^{2/d}p_1}
-\frac{\tr(\nabla\nabla p_2)}{(n_2p_2)^{2/d}p_2},
\end{align*}
where $\nabla\nabla p_1$ and $\nabla\nabla p_2$ 
are the Hessian matrices which are metric-dependent.
Therefore, changing the metric in the input space 
is expected to reduce the bias.

It was shown in \cite{noh2014bias} that the best local Mahalanobis
metric
$(\vector{x}-\vector{x}')^{\top}\widehat{\A}(\vector{x}-\vector{x}')$
for point $\vector{x}$ that minimizes the above approximate bias is
given by
\begin{align*}
\widehat{\A} \propto \left[
            \begin{array}{cc}
              \U_+ & \U_- \\
            \end{array}
          \right]
          \left(
            \begin{array}{cc}
              d_+\Lambda_+ & 0 \\
              0 & -d_-\Lambda_- \\
            \end{array}
          \right)
          \left[
            \begin{array}{cc}
              \U_+ & \U_- \\
            \end{array}
          \right]^\top ,
\end{align*}
where $\Lambda_+\in\R{d_+\times d_+}$ and
$\Lambda_-\in\R{d_-\times d_-}$ are the diagonal matrices containing
$d_+$ positive and $d_-$ negative eigenvalues of $\mathbf{B}$, respectively:
\begin{align*}
     \mathbf{B}&=\frac{1}{((n_1-1)p_1)^{2/d}}
   \frac{\nabla\nabla p_1}{p_1}
   -\frac{1}{(n_2p_2)^{2/d}}\frac{\nabla\nabla p_2}{p_2}.
\end{align*}
The matrices $\widehat{\A}$ and $\mathbf{B}$ share the same eigenvectors,
and $\U_+\in\R{d\times d_+}$ and $\U_-\in\R{d_-\times d_-}$
are collections of eigenvectors that
correspond to the eigenvalues in $\Lambda_+$ and $\Lambda_-$, respectively.

In~\cite{noh2014bias}, the authors assumed that $p_1$ and $p_2$ are both
nearly Gaussian, and estimated densities $p_1$ and $p_2$ as well as their Hessian matrices
$\nabla\nabla p_1$ and $\nabla\nabla p_2$ from the Gaussian models with
maximum likelihood estimation.  It was demonstrated that the accuracy of
NNDE-based KL-divergence approximation is significantly improved when $p_1$ and $p_2$
are nearly Gaussian.

\subsection{Applying MISED to Metric Learning for NNDE-Based KL-Divergence Approximation}
However, the above method does not work well if $p_1$ and $p_2$ are
apart from Gaussian.  Here, we propose to use our non-parametric
density-derivative estimator in metric learning for NNDE-based
KL-divergence approximation.

Since the scale of $\mathbf{B}$ is arbitrary, let us use the following
rescaled matrix $\widetilde{\mathbf{B}}$ instead:
\begin{align}
  \widetilde{\mathbf{B}}&=\frac{1}{(n_1-1)^{2/d}}
  \left\{\frac{p_2}{p_1}\right\}^{2/d+1}\nabla\nabla p_1
  -\frac{1}{n_2^{2/d}}\nabla\nabla p_2.
 \label{B}
\end{align}
We estimate the Hessian matrices $\nabla\nabla p_1$ and $\nabla\nabla
p_2$ by the proposed MISED method, and the density ratio $p_2/p_1$ by the unconstrained
least-squares density-ratio estimator \cite{kanamori2009least} that
directly estimates the density ratio in a non-parametric manner without
estimating each density.  By this, we can perform metric learning in a
non-parametric way without explicitly estimating the densities $p_1$ and
$p_2$.

\begin{figure}[!t]
 \centering \includegraphics[width=\textwidth]{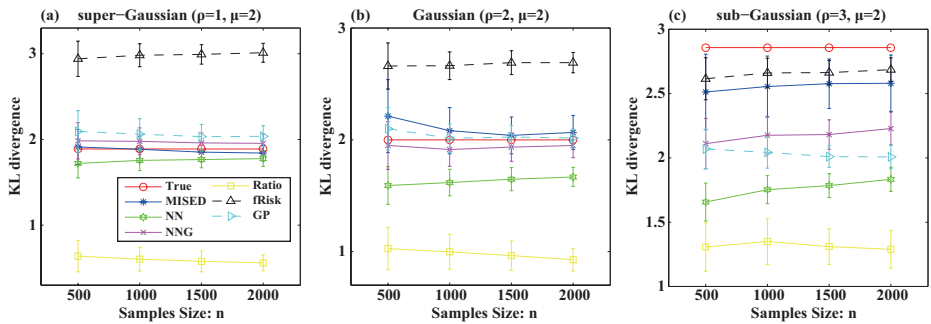}
  \caption{\label{fig:KL} KL-divergence estimation for (a) super-Gaussian,
  (b) Gaussian and (c) sub-Gaussian data as a function of sample size $n$.
}
\end{figure}

\subsection{Numerical Examples}
We experimentally compare the behavior of the NNDE-based KL-divergence
approximator with MISED-based metric learning (MISED),
that without metric learning (NN) \cite{Wang06anearest-neighbor}, 
that with Gaussian-based metric learning (NNG) \cite{noh2014bias},
the density-ratio-based non-parametric KL-divergence estimator (Ratio)
\cite{IEEE-IT:Nguyen+etal:2010},
the risk-based nearest-neighbor KL-divergence estimator (fRisk)
\cite{ICML:Garcia-Garcia+etal:2011}, 
and the Gaussian parametric KL-divergence estimator with maximum likelihood estimation (GP).

We generate data samples from the generalized Gaussian distribution:
\begin{align*}
 p_\mathrm{GG}(x;\mu,\beta,\rho)&=\frac{\beta^{1/2}}{2\Gamma(1+1/\rho)}
 \exp\left(-\beta^{\rho/2}|x-\mu|^{\rho}\right),
\end{align*}
where $\mu\in\mathbb{R}$ denotes the mean, $\beta>0$ controls the variance,
and $\rho>0$ controls the Gaussianity:
$\rho<2$, $\rho=2$, and $\rho>2$ correspond to 
super-Gaussian, Gaussian, and sub-Gaussian distributions, respectively.
For $\vector{x}=(x^{(1)},\ldots,x^{(d)})^\top$ with $d=5$, we set
\begin{align*}
p_1(\vector{x})&=\prod_{j=1}^dp_\mathrm{GG}(x^{(j)};0,\beta,\rho),\\
p_2(\vector{x})&=p_\mathrm{GG}(x^{(1)};2,\beta,\rho)\prod_{j=2}^dp_\mathrm{GG}(x^{(j)};0,\beta,\rho),
\end{align*}
where the value of $\beta$ is selected so that the variance is one.
We evaluate the performance of each method when sample size $n$ and 
Gaussianity $\rho$ are changed.

The experimental results for $\rho=1,2,3$ and $n=500,1000,1500,2000$ are
presented in Figure~\ref{fig:KL}.  The proposed MISED outperforms the
plain NN (without metric learning) for all three cases, and it
outperforms NNG and GP for the super-Gaussian and sub-Gaussian cases.
GP and NNG work the best for the Gaussian case as expected, but MISED
also still works reasonably well. fRisk is better than MISED for
the sub-Gaussian case, but it largely overestimates for the other two
cases. Ratio is a completely non-parametric method, but it
systematically underestimates for all three cases.

\subsection{Experiments on Distributional Change Detection}
\label{ssec:CDec}
\begin{figure}[!t]
 \centering 
 \subfigure[Example 1]{
 \begin{tabular}{@{}c@{}}
  \includegraphics[width=0.47\textwidth]{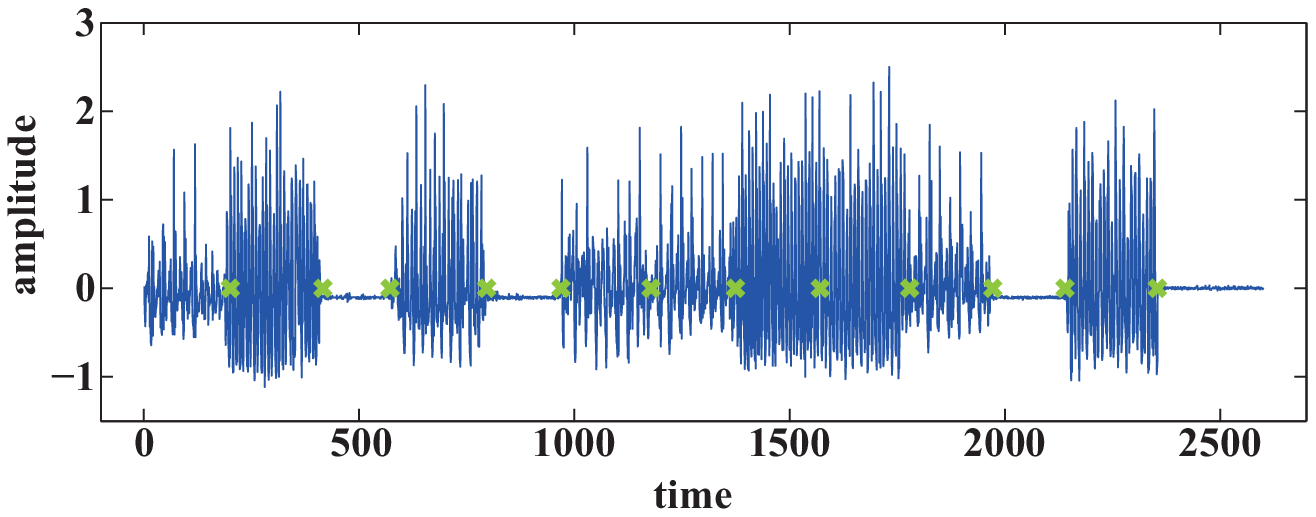}\\
  \includegraphics[width=0.47\textwidth]{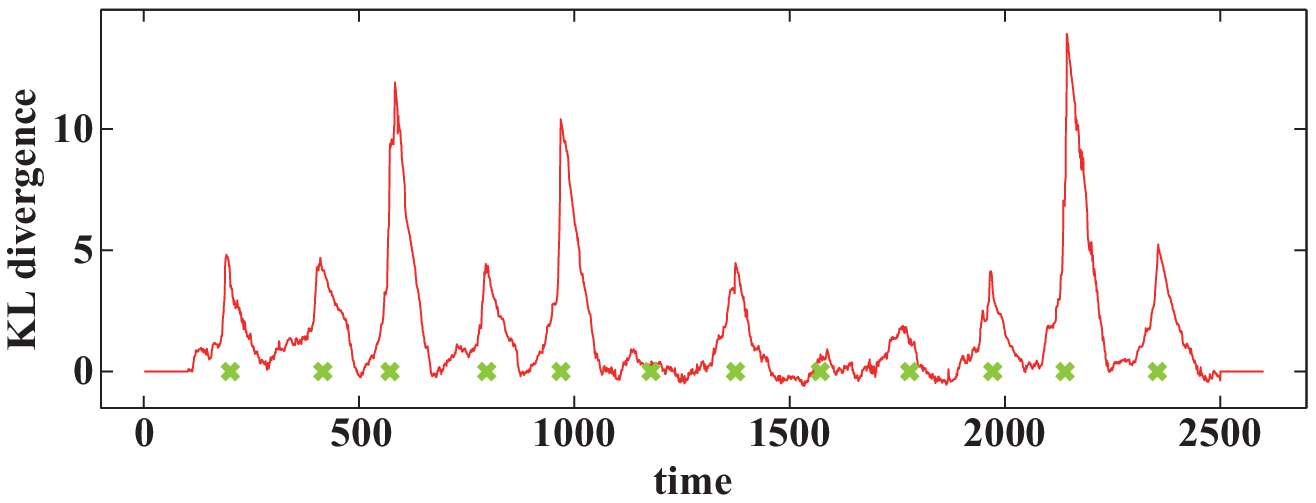}\\
 \end{tabular}
 }
 \subfigure[Example 2]{
 \begin{tabular}{@{}c@{}}
  \includegraphics[width=0.47\textwidth]{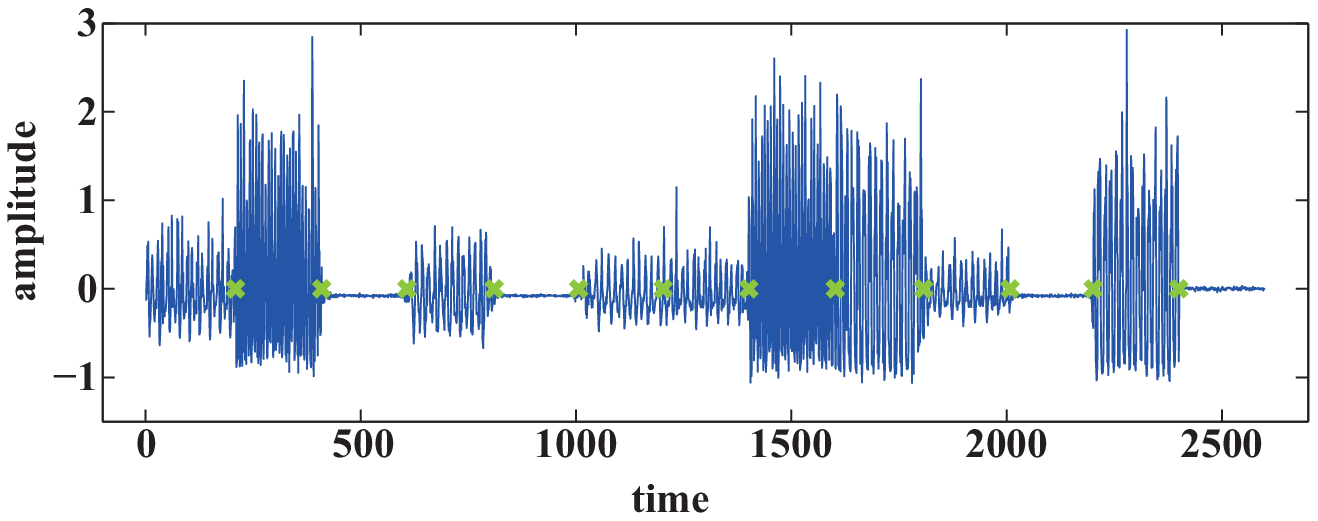}\\
  \includegraphics[width=0.47\textwidth]{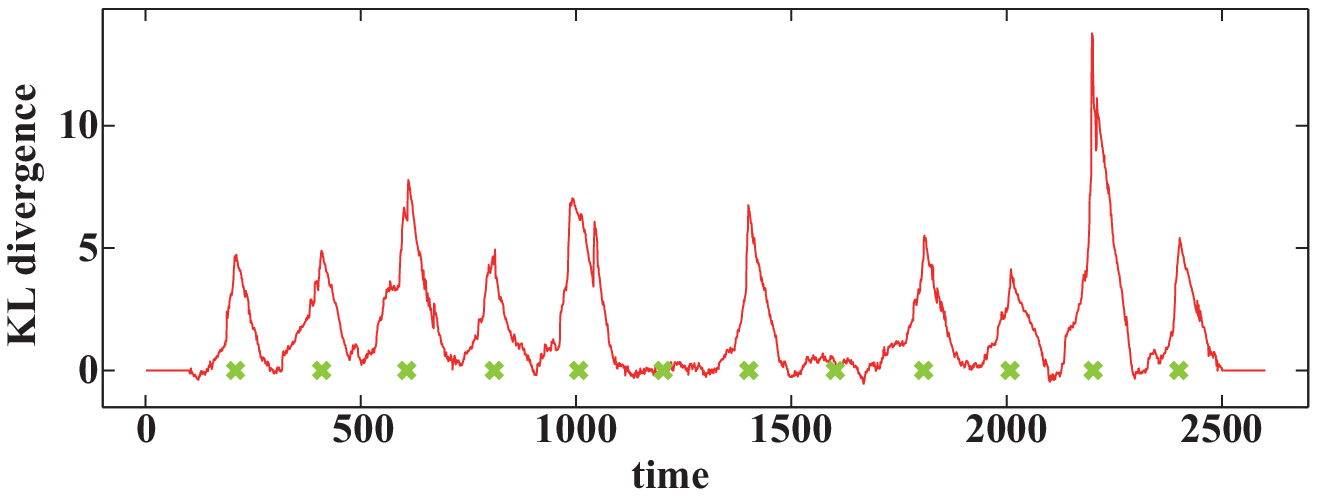}\\
 \end{tabular}
}
 \caption{\label{fig:chdec} HASC time series data (top) and
   the KL-divergence estimated by MISED (bottom).
   Green symbols represent the true change points.}
\end{figure}

\begin{table}[t]
 \caption{\label{tab:AUC} Means and standard deviations of the area
   under the ROC curve (AUC) over $10$ runs.  The best method and
   methods comparable to the best one in terms of the mean AUC by the
   one-tailed Welch's t-test with significance level $5\%$ are
   highlighted in boldface.}  \centering
 \vspace*{2mm}
  \begin{tabular}{c|c|c|c}
   GP & NNG~\cite{noh2014bias} & fRisk~\cite{ICML:Garcia-Garcia+etal:2011} & MISED \\ 
   \hline 
   0.747(0.050) & 0.822(0.030) & {\bf 0.858(0.022)} & {\bf 0.839(0.028)}
  \end{tabular}
\end{table}

The goal of change detection is to find abrupt changes in time-series
data.  We use an $m$-dimensional real vector $\vector{y}(t)$
to represent a segment of time series at time $t$, and a
collection of $r$ such vectors is obtained from a sliding window:
\[
\vector{Y}(t) := \{ \vector{y}(t), \vector{y}(t+1),
\ldots,\vector{y}(t+r-1)\}.
\]
 Following \cite{NN:Liu+etal:2013},
we consider an underlying density function that generates $r$
retrospective vectors in $\vector{Y}(t)$.  We measure the KL-divergence
between the underlying density functions of the two sets,
$\vector{Y}(t)$ and $\vector{Y}(t + r + m)$ for every $t$,
and determine a point $t_0 + r + m$ as a change point if the KL-divergence for
$\vector{Y}(t_0)$ and $\vector{Y}(t_0 + r + m)$ is greater than a
predefined threshold. In the experiment, we set $r=3$ and $m=100$.

We use the \emph{Human Activity Sensing Consortium (HASC) Challenge
2011} collection\footnote{\url{http://hasc.jp/hc2011/}}, which provides human
activity information collected by a portable three-axis
accelerometer. Our task is to segment different activities such as
``stay'', ``walk'', ``jog'', and ``skip''. Because the orientation of
the accelerometer is not necessarily fixed, we took the $\ell_2$-norm of
3-dimensional accelerometer data and obtained one-dimensional data,
following \cite{NN:Liu+etal:2013}.

Figure~\ref{fig:chdec} depicts examples of time-series data and their
KL-divergences (which is regarded as a change score).  These graphs show
that the change scores tend to be large at the true change points.
Next, we more systematically evaluate the performance of change
detection using the AUC (area under the ROC curve) scores.  The results
are summarized in Table~\ref{tab:AUC}, showing that the proposed MISED
outperforms GP and NNG, and is comparable to fRisk.  In the experiments
in Figure~\ref{fig:KL}, fRisk gave similar values for different
distributions even when the true KL-divergence is large.  This was poor
as a KL-divergence approximator, but this property seems to work as a
``regularizer'' to stabilize the change score to avoid incurring big
error.  Similar tendencies were also reported in the previous work
\cite{noh2014bias}.

\subsection{Experiments on Information-Theoretic Feature Selection}
Finally, KL-divergence approximation is applied to selecting relevant
features for classification.  The \emph{Jensen-Shannon (JS) divergence}
is an information-theoretic measure between labels $y\in\{1,2\}$ and
features $\vector{x}\in\mathbb{R}^d$:
\begin{align*}
\mathrm{JS}(\mathcal{X};\vector{y}) 
&= -\sum_{y = 1}^2\int p(\vector{x},y)\log\frac{p(\vector{x})p(y)}{p(\vector{x},y)}
\mathrm{d}\vector{x}
\\
&=
p(y=1) \mathrm{KL}(p(\vector{x}|y=1)\|p(\vector{x}))\\
&\phantom{=}+ p(y=2) \mathrm{KL}(p(\vector{x}|y=2)\|p(\vector{x})),
\end{align*}
where $p(\vector{x}) = p(y=1) p(\vector{x}|y=1) + p(y=2)
p(\vector{x}|y=2)$.

\begin{figure}[!t] \centering \includegraphics[width=3.5in]{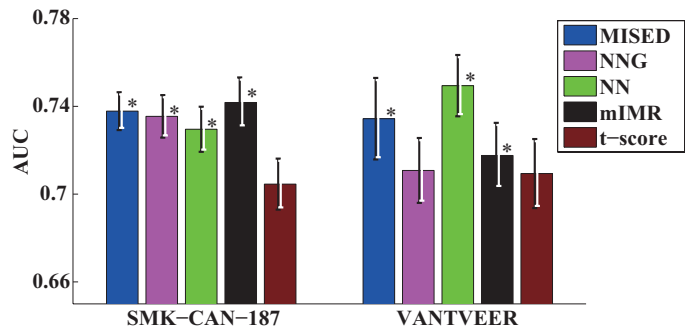}
\caption{Gene expression classification with feature selection. The best
method and methods comparable to the best one in terms of the mean AUC
by the one-tailed Welch's t-test with significance level $5\%$ are
highlighted by the asterisks.}  \label{fig:geneSelection}
\end{figure}

We use two gene expression datasets of breast cancer prognosis studies:
``SMK-CAN-187'' \cite{Freije2004Gene} and ``VANTVEER''
\cite{Spira2007Airway}.  The SMK-CAN-187 dataset contains 90 positive
(alive) and 97 negative (dead after 5 years) samples with 19993
features.  We use 65 randomly selected samples per class for training
and use the rest for evaluating the test classification performance.
The VANTVEER dataset contains 46 positive and 51 negative samples with
24481 features. We use 35 randomly selected data per class for training
and use the rest for evaluating the test classification performance.

We choose $20$ features based on the forward selection strategy and
compare the AUC of classification.  The results are summarized in
Figure~\ref{fig:geneSelection}, showing that the proposed method works
reasonably well.

\section{Conclusion}
We proposed a method to directly estimate density derivatives. The
proposed estimator, called MISED, was shown to possess various useful
properties, e.g., analytic and computationally efficient estimation
of multi-dimensional high-order density derivatives is possible and all
hyper-parameters can be chosen objectively by cross-validation.  We
further proposed a MISED-based metric learning method to improve the
accuracy of nearest-neighbor KL-divergence approximation, and its practical
usefulness was experimentally demonstrated on change detection and
feature selection.

Estimation of density derivatives is versatile and useful in
various machine learning tasks beyond KL-divergence approximation.
In our future work, we will explore more applications
based on the proposed MISED method.

\bibliographystyle{unsrt}
\bibliography{papers.bib}

\end{document}